%% file: paper.tex
\documentclass[conference]{IEEEtran}
\IEEEoverridecommandlockouts

\usepackage{cite}
\usepackage{amsmath,amssymb,amsfonts}
\usepackage{algorithmic}
\usepackage{graphicx}
\usepackage{textcomp}
\usepackage{xcolor}
\usepackage{algorithm}
\usepackage{algorithmic}
\usepackage{todonotes}
\usepackage[para,multiple]{footmisc}

\newcommand{\data}{{\sffamily \small{Data}}}
\newcommand{\shard}{{\sffamily \small{Shard}}}
\newcommand{\pipeshard}{{\sffamily \small{Pipeshard}}}
\newcommand{\zero}{{\sffamily \small{ZeRO2}}}
\newcommand{\gptm}{{\it gpt2m}}
\newcommand{\gptl}{{\it gpt2l}}
\newcommand{\gptL}{{\it gpt2L}}

\def\BibTeX{{\rm B\kern-.05em{\sc i\kern-.025em b}\kern-.08em
    T\kern-.1667em\lower.7ex\hbox{E}\kern-.125emX}}
\begin{document}

\title{Performance of Small Language Model Pretraining on FABRIC: An Empirical Study}	

\author{\IEEEauthorblockN{Praveen Rao}
\IEEEauthorblockA{
\textit{The University of Missouri}\\
Columbia, USA \\
praveen.rao@missouri.edu}
}

\maketitle

\input{abstract}


\input{intro}
\input{background}
\input{approach}
\input{experiments}

\input{conclusion}

\bibliographystyle{IEEEtran}
\bibliography{ref-rao,rdf}
\end{document}

%% file: abstract.tex
\begin{abstract}
Large language models (LLMs) require enormous computing power to pretrain on massive datasets. When limited datasets are available, smaller-sized LLMs are better choice to pretrain (on user-specified datasets) by following the scaling laws of LLMs. Using pretrained models, vector embeddings can be generated for raw data and stored using vector databases to support modern AI applications and semantic search. In this work, we investigate the performance of pretraining techniques for smaller-sized LLMs on an experimental testbed (with commodity GPUs) available to academic users at no charge. We consider data parallelism, intra-operator parallelism, and inter-operator/pipeline parallelism, and their combinations for pretraining. We set up different GPU clusters with homogeneous and heterogeneous GPU hardware. Furthermore, we investigate the impact of network latency on pretraining performance especially when GPUs are geographically distributed. We used GPT-2 medium and large models and pretrained them using open-source packages, namely, Alpa and Ray. We observed that Alpa's execution plans that collectively optimized intra-operator and inter-operator/pipeline parallelism consistently performed the best when GPUs were geographically distributed. This was especially true when the network latencies were in 10's of milliseconds. Based on the insights gained from the experiments, we propose a systematic approach for selecting the appropriate pretraining technique to achieve high training performance/lower execution time as well as to reduce the number of GPUs used.
\end{abstract}



%% file: intro.tex
\section{Introduction}

Artificial intelligence (AI) has become ubiquitous and drives a plethora of real-world applications in information retrieval (IR), natural language processing (NLP), computer vision, speech recognition, e-commerce, healthcare, and  defense. Generative AI services (e.g., ChatGPT~\cite{ChatGPT}, DALL.E~\cite{DallE}), which can generate new content of different modalities such as text and images, have had an explosive growth in recent years~\cite{McKinsey2024}. It is predicted that generative AI will become a \$1.3 trillion market by 2032~\cite{Bloomberg2023}. Large language models (LLMs) (a.k.a. foundation models), which are trained on a large corpus of unlabeled data via self-supervised learning~\cite{ericsson2022self}, have become the bedrock of generative AI. Several LLMs have been proposed in recent years, namely, GPT-3/GPT-4~\cite{GPT32020,openai2024gpt4technicalreport}, Jurassic~\cite{lieber2021jurassic}, Gopher~\cite{Gopher2023}, Megatron-Turing NLG 530B~\cite{MTNLG2022}, OPT~\cite{zhang2022opt}, and LaMDA~\cite{LaMDA2022}. Google's PaLM~\cite{PaLM}/MedGemma~\cite{sellergren2025medgemmatechnicalreport}, Meta's Llama~\cite{LLaMA}, NVIDIA's NVLM~\cite{dai2024nvlm}, Anthropic's Claude~\cite{Claude}, Alibaba's Qwen~\cite{bai2023qwentechnicalreport}, Grok3~\cite{Grok3}, and DeepSeek~\cite{deepseekai2025deepseekv3technicalreport} family of models are other notable contributions~\cite{zhao2023surveylargelanguagemodels,minaee2024largelanguagemodelssurvey}. Indeed, LLMs have taken over the world of AI like a storm.


At the heart of an LLM is the Transformer architecture proposed by Vaswani et. al. in 2017~\cite{vaswani2017attention}. A Transformer relies on the idea of self-attention and trains efficiently due to parallelizable components using self-supervised learning. Given a sequence of tokens, it learns to predict the next token, and hence, is an autoregressive model. Since its inception, researchers have applied Transformers-based models to numerous domains such as NLP, IR, image classification, question answering, and so on~\cite{zhao2023surveylargelanguagemodels,minaee2024largelanguagemodelssurvey,ren2023rejuvenating,chen2020generative,he2023transformers,ViT2020}.

Pretraining is a type of \textit{self-supervised training} performed on large datasets (e.g., autoregressively to predict the next token, causal language modeling); the model weights of an LLM can be randomly initialized for pretraining. LLMs pretrained on massive datasets use 100's of GPUs~\cite{Anyscale1000GPUs} costing millions of dollars. It is claimed that OpenAI's GPT-3 (with 175 billion parameters) cost more than \$4.6 million to train~\cite{GPT3Cost}. Databricks spent \$10 million to build their LLM called DBRX~\cite{DBRXCost}. Meta used clusters with 24K GPUs to train Llama3~\cite{MetaLLMTraining}. More recently, it was estimated that the cost of training DeepSeek-R1 was \$294K~\cite{guo2025deepseek}. To accelerate LLM training, researchers have explored quantization techniques to reduce the model size such as using 8-bit floating point (FP8)~\cite{micikevicius2022fp8formatsdeeplearning,deepseekai2025deepseekv3technicalreport,mishra2025recipespretrainingllmsmxfp8} and 4-bit floating point (FP4)~\cite{wang2025optimizinglargelanguagemodel,pmlr-v258-tseng25a,nvidia2025pretraininglargelanguagemodels,zhou2025efficientpretrainingexploringfp4}.

While the race for building larger and more capable LLMs continues to drive technology companies, there is a growing need in academia to empower users to pretrain language models to foster new scientific discoveries and accelerate innovation. While is it impossible to level the playing field for academic users in terms of access to massive compute resources/datasets for training LLMs, we posit that \textit{federally funded testbeds} could be leveraged in a creative way if we shift the focus to pretraining small language models (SLMs). 

The motivation for pretraining SLMs on user-specified datasets is threefold: Firstly, scaling laws for LLMs~\cite{ScalingLawsLLM,hoffmann2022training} have provided insights into the relationship between model size, dataset size, and compute budget. For compute-optimal training, Hoffmann et. al.~\cite{hoffmann2022training} showed that if the model size is doubled, then the number of training tokens should also be doubled. Their compute-optimal model \textit{Chinchilla} with 70 billion parameters outperformed models with 300+ billion parameters. Overfitting can arise if the dataset size is fixed but the model size is increased~\cite{ScalingLawsLLM}. As academic users usually have smaller datasets compared to industry teams and cannot afford a huge compute budget, it is not pragmatic to pretrain LLMs (e.g., with 100-500 billion parameters). \textit{We believe that pretraining SLMs} (e.g., GPT-2~\cite{Radford2019LanguageMA}, GPT-3-13B~\cite{GPT32020}, OPT7B~\cite{zhang2022opt}, Llama3-33B~\cite{touvron2023llama}, Mistral7B~\cite{jiang2023mistral}) \textit{is an apt choice for academic use cases}. In fact, there has been growing interest in pretraining SLMs on commodity GPUs~\cite{Sanyal2024pretraining,sanyal2025,SmallLLM2024}. Further, SLMs are touted as the future of agentic AI~\cite{belcak2025smalllanguagemodelsfuture}.



Secondly, domain-specific LLMs~\cite{Patel2024} (e.g., LEGAL-BERT~\cite{chalkidis-etal-2020-legal} for law, Med-PaLM~\cite{singhal2023large}/MedGemma~\cite{sellergren2025medgemmatechnicalreport} for medicine, BloombergGPT for finance~\cite{wu2023bloomberggpt}) are becoming more relevant as generalist LLMs lack sufficient domain expertise, can hallucinate, and be unaware of current events leading to AI mistakes/failures~\cite{AIfailures}. Hence, it is appealing for users to pretrain SLMs on specialized datasets containing domain-specific knowledge to begin with. While fine-tuning of generalist LLMs can make them domain-specific, the process is reported to be time consuming and expensive for LLMs~\cite{Patel2024}.

Thirdly, vector databases\footnote{www.pinecone.io}\footnote{www.trychroma.com}\footnote{https://faiss.ai}\footnote{https://qdrant.tech} are becoming popular in AI applications, semantic search, and retrieval-augmented generation (RAG) systems. Interestingly, data management systems such as MongoDB, Postgres, Apache Solr, Redis, and Apache Cassandra have extended their support for vector search. Generating dense numerical vectors or embeddings is the fundamental task for representing different modalities of data such as text and images. Using $k$-nearest neighbor search on embeddings, similar data can be obtained for a given input. Pretrained SLMs can be used to generate embeddings of raw data. However, generating accurate embeddings in domain-specific applications (e.g., healthcare, life sciences, engineering, defense, education) demands SLMs be pretrained on user-specified datasets.



Motivated by the aforementioned reasons, we investigate the performance of different SLM pretraining techniques on FABRIC~\cite{FABRIC2023}, an NSF-funded nationwide research infrastructure. This infrastructure is available \textit{at no charge} to academic users. Our ultimate goal is to democratize SLM pretraining (and inference) for academic users using the capabilities of FABRIC to build domain-specific vector databases. In this paper, we make the following key contributions:

\begin{itemize}
\item We studied the performance of different parallelization techniques for pretraining SLMs using commodity GPUs on FABRIC. We consider data parallelism, intra-operator, inter-operator/pipeline parallelism, and their combinations. 
\item We evaluate the SLM pretraining performance of GPT-2 medium and large models using open-source packages, namely, Alpa~\cite{zheng2022alpa} and Ray~\cite{moritz2018ray}. We leverage different types of GPU clusters on FABRIC. We explore how network latencies between GPUs (due to being geographically distributed) impact training performance. Both homogeneous and heterogeneous GPU hardware are considered.
\item Based on the evaluation, we observed that Alpa’s pretraining execution plans that collectively optimized intra-operator and inter-operator/pipeline parallelism consistently performed the best when GPUs were geographically distributed with network latencies in 10’s of milliseconds. However, Alpa's plans ran slower than techniques like data parallelism when an SLM could be pretrained on GPUs attached to a single virtual machine (VM).
\item Based on the insights gained from the evaluation, we propose a systematic approach for selecting the appropriate pretraining technique to achieve \textit{high} training performance and \textit{lower} total training time as well as to \textit{reduce} the number of GPUs used.
\end{itemize}

The rest of the paper is organized as follows: Section~\ref{sec-background} discusses background and related work. Section~\ref{sec-approach} introduces our methodology for the evaluation. Section~\ref{sec-evaluation} presents the experimental results and key findings. Finally, we conclude in Section~\ref{sec-conclusion}.

%% file: background.tex
\section{Background and Related Work}
\label{sec-background}


\subsection{FABRIC}

FABRIC~\cite{FABRIC2023} is a programmable infrastructure with extensible networking elements and large amounts of compute and storage capabilities scattered throughout the network. A FABRIC node is equipped with powerful processors, large amounts of RAM, non-volatile memory express (NVMe) drives, GPUs, field-programmable gate arrays (FPGAs), 100/200 Gbps network interface cards (NICs), and programmable SmartNICs~\cite{SmartNICs}. Spanning North America and Europe, FABRIC has 39 sites totaling 88,192 cores, 1.94 PB of disk storage, 61 TB RAM, 173 GPUs, 157 SmartNICs, 370 NVMe drives, and 23 FPGAs. High speed optical links (e.g., Terabit Core, 100 Gbps Layer 1/2) interconnect different sites. An experimenter can create a \textit{slice} (or an experiment) containing several VMs to form a cluster. They can attach GPUs, SmartNICs, and NVMe drives to each VM in a programmable way. The slice can span a single FABRIC site (e.g., Utah) or multiple FABRIC sites (e.g., Utah $\leftrightarrow$ Dallas $\leftrightarrow$ Amsterdam). (Each site can be thought of as a mini data center.)

FABRIC offers specialized network services to connect VMs~\cite{FABRIC-NW}. The Layer 2 Bridge Service (L2Bridge) is designed to connect VMs in a single site using Layer 2 broadcasting based on virtual local area networks (LANs). The Layer 2 Site-to-Site Connection Service (L2STS) is designed to connect VMs between two different FABRIC sites. L2STS is based on Ethernet Virtual Private Networking. FABRIC also supports Layer 3 network services.

\subsection{Distributed Systems for AI Workloads}

Several systems have been developed for large-scale machine/deep learning. Horovod~\cite{sergeev2018horovod}, PyTorch~\cite{paszke2019pytorch}, TensorFlow~\cite{abadi2016tensorflow}, and MXNet~\cite{chen2015mxnet} enable distributed training. However, they do not support fine-grained simulation. Hence, Ray~\cite{moritz2018ray} was developed to support scalable reinforcement learning. It expresses task-parallel and actor-based computations using a single interface and supports asynchronous tasks. For scalability and fault tolerance, it employs a distributed scheduler and stores the control state in a distributed manner. Today, Ray~\cite{Ray} is an open-source distributed system for scaling AI workloads involving deep learning and machine learning. It is used by 10,000+ organizations including OpenAI, Uber, AWS, Netflix, Instacart, and many others. 


Horovod~\cite{sergeev2018horovod} and PyTorchDDP~\cite{li2020pytorch} were early systems that employed data parallelism for distributed deep learning. Subsequently, ZeRO~\cite{rajbhandari2020zero} improved the memory usage of data parallelism for large models via optimizer state memory optimization. ZeRO-2~\cite{ZeRO2} further enhanced the optimizations of ZeRO by reducing the memory footprint of gradients, activation memory, and fragmented memory. Model parallelism is needed when the size of a deep learning model is larger than a single GPU's memory. The deep learning training process of a model can be represented as \textit{a dataflow computational graph}~\cite{bradbury2018jax,abadi2016tensorflow,paszke2019pytorch} where the nodes are operators such as matrix multiplication that operate on tensors. Both inter-operation and intra-operator parallelism can be employed. Several systems allow users to provide manual parallel plans~\cite{shazeer2018mesh,xu2021gspmd,huang2019gpipe,rasley2020deepspeed}. For Transformer models, Megatron-LM~\cite{shoeybi2019megatron,narayanan2021efficient,MTNLG2022,korthikanti2023reducing} and TeraPipe~\cite{li2021terapipe} were developed that provided specialized partitioning strategies. StellaTrain~\cite{lim2024accelerating} introduced acceleration techniques to maximize GPU utilization for distributed training; it achieved significant improvement over PyTorch DDP in a multi-cluster setting with low wide-area network (WAN) bandwidth between clusters.
 

More recent systems automatically search for the best parallel plans~\cite{zheng2022alpa,unger2022unity,jia2019beyond,zhang2020autosync}. Alpa~\cite{zheng2022alpa} automatically creates model parallel plans by combining data, operator, and pipeline parallelisms. Both \textit{intra-operator} and \textit{inter-operator} parallelism are employed in the plans. Alpa partitions the cluster into a number of device meshes. Each mesh is two dimensional with homogeneous GPUs (same compute capability) and can communicate with higher bandwidth along the first dimension (e.g., on the same machine) compared to the second (e.g., via Ethernet). \textit{The computational graph} is partitioned into stages. Each stage is assigned to a mesh and executes using intra-operator parallelism. Collective communication~\cite{chan2007collective} is done within a mesh. Inter-operator parallelism occurs between meshes by splitting the training batch into microbatches and pipelining the forward and backward passes across the microbatches. For inter-operator parallelism, point-to-point communication is performed between different hosts/devices. To improve the communication efficiency during training, recently cross-mesh sharding was introduced in Alpa~\cite{zhuang2023optimizing}. Anyscale~\cite{Anyscale1000GPUs} showed that Alpa and Ray can be used to train a 175B LLM using 1000 GPUs. 

\subsection{Caveats Using FABRIC for Pretraining}

There are several caveats to consider when pretraining SLMs on FABRIC. The first is the limited number of GPUs available at each FABRIC site. This limits the number of GPUs that can be attached to a single VM. Hence, VMs must be stitched together to increase the number of GPUs to form a cluster. Therefore, different sites on FABRIC should be involved. Second, a cluster that is provisioned may have heterogeneous GPU hardware as other academic users are also utilizing resources on FABRIC. Getting all the GPUs of the same type for an experiment may not be possible. Third, the site-to-site network latencies in FABRIC is non-trivial and can vary in the range of 10's of milliseconds. This can significantly increase the cost of communication between GPUs during pretraining. Furthermore, the collective communication between GPUs on different VMs use TCP/IP instead of direct GPU-to-GPU interconnects\footnote{https://www.nvidia.com/en-us/data-center/nvlink}.

%% file: approach.tex
\section{Methodology}
\label{sec-approach}

In this section, we present our experimental methodology and discuss different parallelization techniques for SLM pretraining that were evaluated on FABRIC~\cite{FABRIC}. 


\subsection{Parallelization Techniques Considered}

\begin{figure}[tbh]
\begin{center}
\includegraphics[width=3.5in, angle=0]{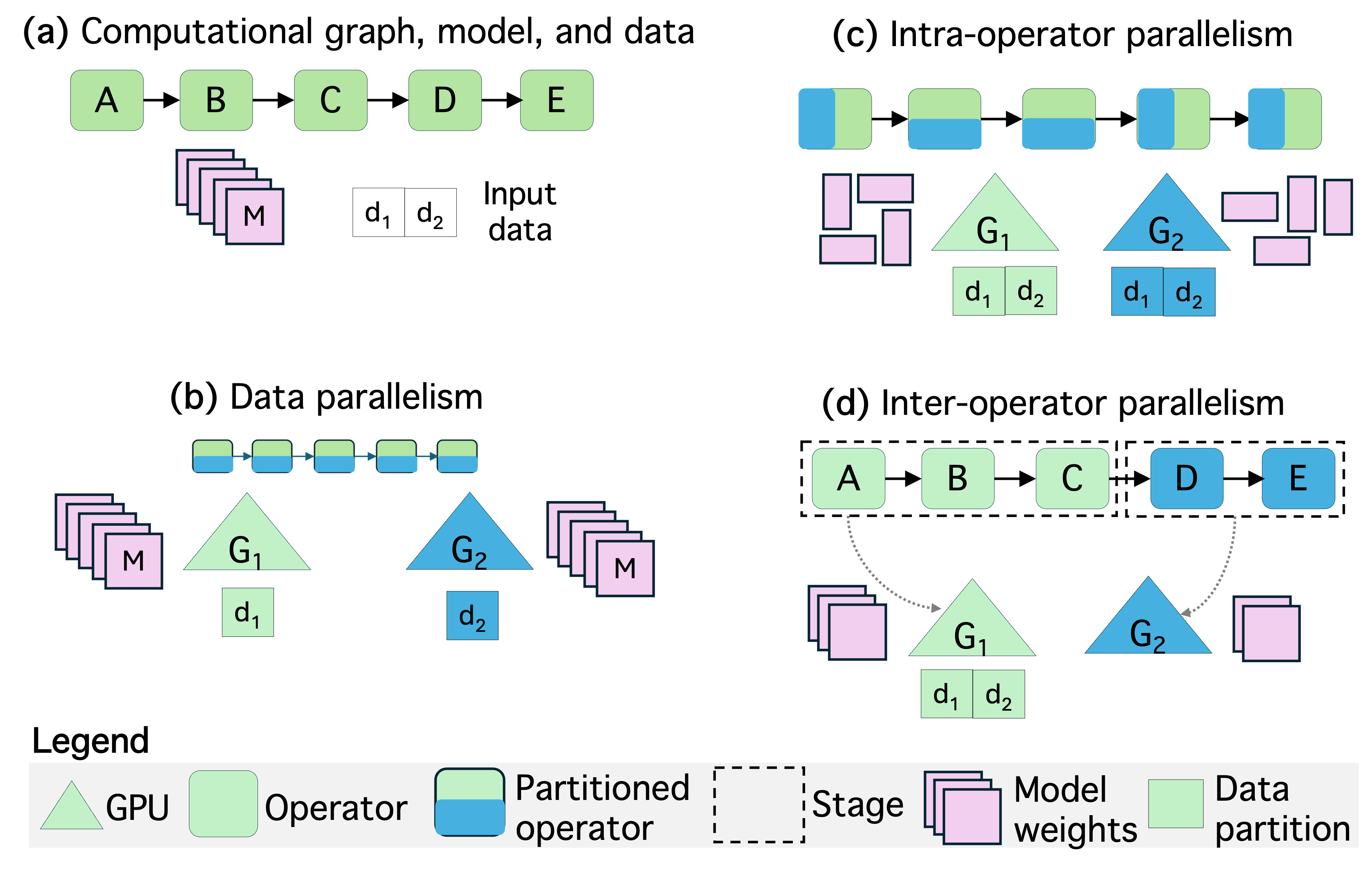}
\end{center}
\caption{Examples of different parallelization techniques}
\label{fig-parallel-techniques}
\end{figure}

\paragraph{Data, Intra-Operator, and Inter-Operator Parallelism} In data parallelism, a model is replicated on each GPU and trained on different splits of the input data. The gradients are synchronized across the GPUs before the weight updates. Consider a computational graph for a model $M$ shown in Figure~\ref{fig-parallel-techniques}(a). (We show only the forward pass for ease of exposition.) Figure~\ref{fig-parallel-techniques}(b) shows the replication of the model on the GPUs and splits of the input data. Each operator of the computation graph is computed across the two GPUs. 

When a model is too large to fit in a single GPU's memory, model parallelism can be used via operator parallelism. In intra-operator parallelism, the multidimensional tensors are partitioned on some dimension and stored on different GPUs; the operators are computed in a distributed manner. Collective communication is required to ensure the right partitions are available to the GPUs. An example is shown in Figure~\ref{fig-parallel-techniques}(c). In fact, data parallelism is a type of intra-operator parallelism. Lastly, in inter-operator parallelism (or pipeline parallelism), the computation graph is split in stages, and each stage is executed on a different GPU. The training batch is split into microbatches; the forward and backward passes are pipelined across the microbatches. Point-to-point communication is needed between GPUs/hosts. An example is shown in Figure~\ref{fig-parallel-techniques}(d).

\paragraph{Alpa's Shard and Pipeshard Parallelism}

Alpa~\cite{zheng2022alpa} explores a hierarchical space of executions plans that combine intra-operator parallelism (or shard parallelism) and inter-operator parallelism (or pipeline parallelism) for pretraining. Alpa's single program, multiple data (SPMD)-style shard/intra-operator parallelism includes data parallelism, operator parallelism, ZeRO optimizer, and their combinations~\cite{Alpaproject}. Integer linear programming is used to minimize the execution cost. Alpa's inter-operator/pipeline parallelism aims to minimize the end-to-end pipeline execution latency for the entire graph via dynamic programming. It identifies the best assignment of stages to GPU meshes by exploring the intra-operator plans for each stage. A $(n\times m)$ mesh is a 2D array of GPUs containing $n$ rows of $m$ GPUs. Alpa makes assumptions about the GPU cluster; each mesh is assumed to have similar capacity. High bandwidth is available between GPUs in a mesh. Alpa also does not model the communication cost between different stages to simplify the process of solving the optimization problem.








\subsection{Pretraining Techniques, Software, Dataset and Models}

Hereinafter, we refer to the different parallelization techniques considered for evaluation as \data{} (data parallelism on one or more VMs), \zero~\cite{ZeRO2}, \shard{} (includes only Alpa's shard parallelism), and \pipeshard{} (combines Alpa's pipeline and shard parallelism). They were compared by measuring the total wall-clock time for pretraining as well as the average training performance (in TFLOP/s) achieved during pretraining. (We followed the same approach used by the authors of Alpa~\cite{zheng2022alpa}.)

We used the original code of Alpa available on GitHub~\cite{Alpaproject}. We rebuilt the code from scratch as the Alpa repository is now archived and read-only. We used Python 3.8.10, CUDA 11.8, CuPy, NCCL 2.15.1, cuDNN 8.8.0, GCC/G++ 7.5.0, Protobuf 3.20.3, and gRPC Python 1.43.0 to build Alpa on Ubuntu Linux 20.04. Ray 2.1.0 was used to manage the cluster CPUs/GPUs/RAM and run the execution plans generated by Alpa for pretraining.

We used the Wikipedia dataset available on HuggingFace~\cite{wikidump}. We selected \textit{20231101.ace}, which is a modest-sized file yet good enough to test the pretraining performance. We considered GPT-2 models~\cite{Radford2019LanguageMA} that are already supported by Alpa. We tested the medium and large GPT-2 models. The GPT-2 medium model (\gptm) had $n\_ctx = 1024$, $n\_embd = 1024$, $n\_head = 16$, $n\_layer = 24$, and $n\_positions = 1024$. The GPT-2 large model (\gptL) had $n\_ctx = 1024$, $n\_embd = 1280$, $n\_head = 20$, $n\_layer = 30$, and $n\_positions = 1024$. In some cases, we had to set $n\_layer=26$ when GPU memory was insufficient for \gptL. We refer to this modified model as \gptl. We ran all the pretraining tasks for \textit{20 epochs}.



\subsection{GPU Cluster Setup}


\begin{figure}[t]
\begin{center}
\includegraphics[width=3.6in, angle=0]{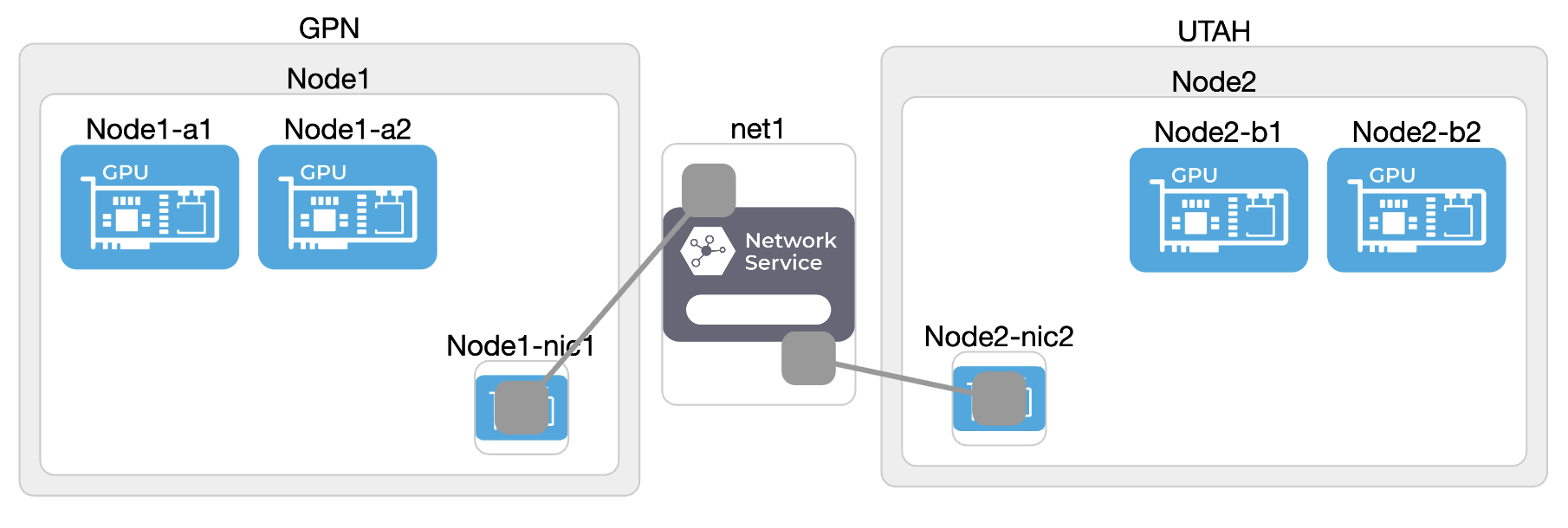}
\end{center}
\caption{A FABRIC slice consisting of two VMs; each VM has 2 GPUs attached to it; the VMs are connected by L2STS}
\label{fig-example-slice}
\end{figure}

On FABRIC, we set up five slices/experiments with different GPU cluster configurations. Each cluster consisted of two VMs with 2 GPUs attached to each VM. Each VM had 12 cores, 32 GB RAM, and 500 GB storage and was allocated in a single FABRIC site. The VMs were connected using L2Bridge or L2STS depending on whether the VMs were in the same site or spanned two different sites (i.e., geographically distributed). Figure~\ref{fig-example-slice} shows an example of a slice spanning two FABRIC sites; the VMs were connected using L2STS. While NCCL (NVIDIA Collective Communication Library) was configured for GPU-GPU communication, it used TCP/IP (IPv4/IPv6) for communication between GPUs on different VMs. 

Three types of GPUs were used in our evaluation, namely, NVIDIA Quadro RTX 6000 (Turing) with 24GB VRAM, NVIDIA Tesla T4 with 16GB VRAM, and NVIDIA A30 (Ampere) with 24GB VRAM. We refer to them as \texttt{RTX}, \texttt{T4}, and \texttt{A30}, respectively in the rest of the discussion. T4 was the least powerful. While RTX and A30 had the same VRAM, A30 was better suited for AI workloads due to its faster memory bandwidth. The details of the five slices are shown in Table~\ref{table-FABRIC-slices}. We show the GPU types used and the site-to-site latency between the VMs (measured using \texttt{ping}). For example, TACC-TACC denotes a \textit{single-site} experiment in Texas. The remaining experiments are \textit{two-site} experiments spanning two different FABRIC sites. Note that the site UTAH was in Utah; GPN was in Missouri; BRIS was in Bristol, UK; GAT was in Georgia; AMST was in Amsterdam, Netherlands; STAR was in Michigan; and MASS was in Massachusetts.

\begin{table}[t]
\caption{FABRIC slice/experiment configurations}
\label{table-FABRIC-slices}
\begin{center}
\begin{tabular}{|c|c|l|l|l|r|}
\hline
\multicolumn{1}{|c|}{\textbf{FABRIC}} &
\multicolumn{1}{|c|}{\textbf{\# of}} &
\multicolumn{1}{|c|}{\textbf{Sites 1}} &
\multicolumn{1}{|c|}{\textbf{Sites 2}} &
\multicolumn{1}{|c|}{\textbf{Network}} &
\multicolumn{1}{|c|}{\textbf{Site-Site}} \\
\multicolumn{1}{|c|}{\textbf{Site 1-Site 2}} &
\multicolumn{1}{|c|}{\textbf{VMs}} &
\multicolumn{1}{|c|}{\textbf{GPUs}} &
\multicolumn{1}{|c|}{\textbf{GPUs}} &
\multicolumn{1}{|c|}{\textbf{Service}} &
\multicolumn{1}{|c|}{\textbf{Latency}} \\
\multicolumn{1}{|c|}{\textbf{}} &
\multicolumn{1}{|c|}{\textbf{}} &
\multicolumn{1}{|c|}{\textbf{}} &
\multicolumn{1}{|c|}{\textbf{}} &
\multicolumn{1}{|c|}{\textbf{}} &
\multicolumn{1}{|c|}{\textbf{(ms)}} \\
\hline
TACC$^\textbf{\dag}$-TACC$^\textbf{\dag}$ & 2 & 2 RTX & 2 T4 & L2Bridge & 0.1\\
\hline
UTAH$^\textbf{\dag}$-GPN$^\textbf{\dag}$ & 2 & 2 RTX & 2 T4 & L2STS & 20.2\\
\hline
UTAH$^\textbf{\dag}$-MASS$^\textbf{\dag}$ & 2 & 2 RTX & 2 RTX & L2STS & 57.4\\
\hline
BRIS$^\textbf{\S}$-STAR$^\textbf{\dag}$ & 2 & 2 A30 & 2 RTX & L2STS & 95.9\\
\hline
GAT$^\textbf{\dag}$-AMST$^\textbf{\S}$ & 2 & 2 A30 & 2 A30 & L2STS & 103.0\\
\hline
\multicolumn{6}{l}{$^\textbf{\dag}$ indicates North America; $^\textbf{\S}$ indicates Europe}
\end{tabular}    
\end{center}
\end{table}

To test the use of all the GPUs in a slice for pretraining, the Ray server was executed on one VM, and a Ray worker was executed on the other. To test the use of only 2 GPUs on a single VM for pretraining, only the Ray server was run. This way the same slice could be used for multiple experiments. 



%% file: experiments.tex
\section{Performance Evaluation}
\label{sec-evaluation}

In this section, we present the performance evaluation results for different pretraining techniques on the five FABRIC slices. Based on the insights gained from the evaluation, we will present a systematic approach for selecting the appropriate pretraining technique.

\subsection{TACC-TACC}

\begin{figure}[t]
\begin{center}
\begin{tabular}{c}
\includegraphics[width=3.6in, angle=0]{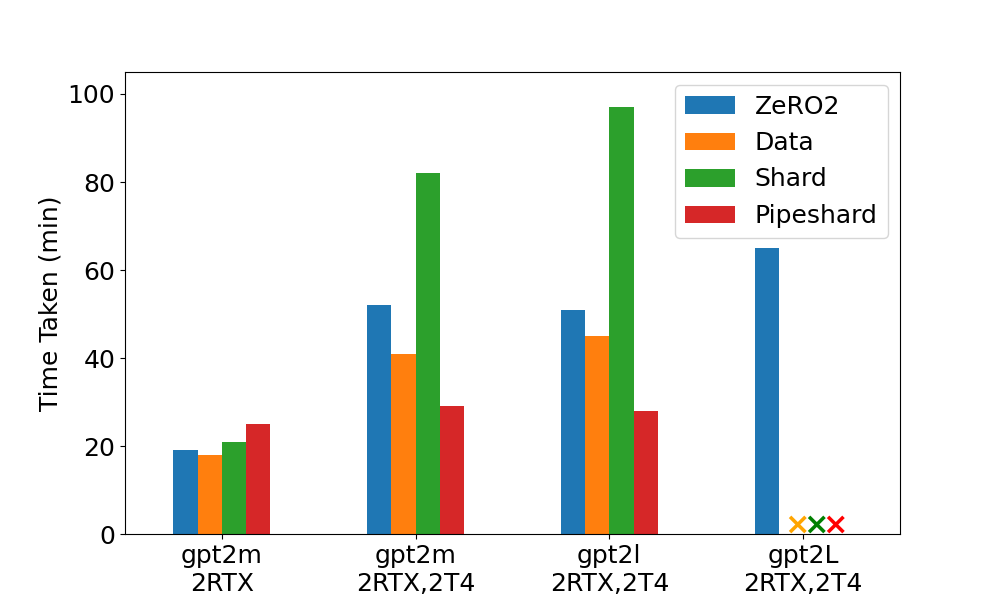} \\
(a) Execution time\\
\includegraphics[width=3.6in, angle=0]{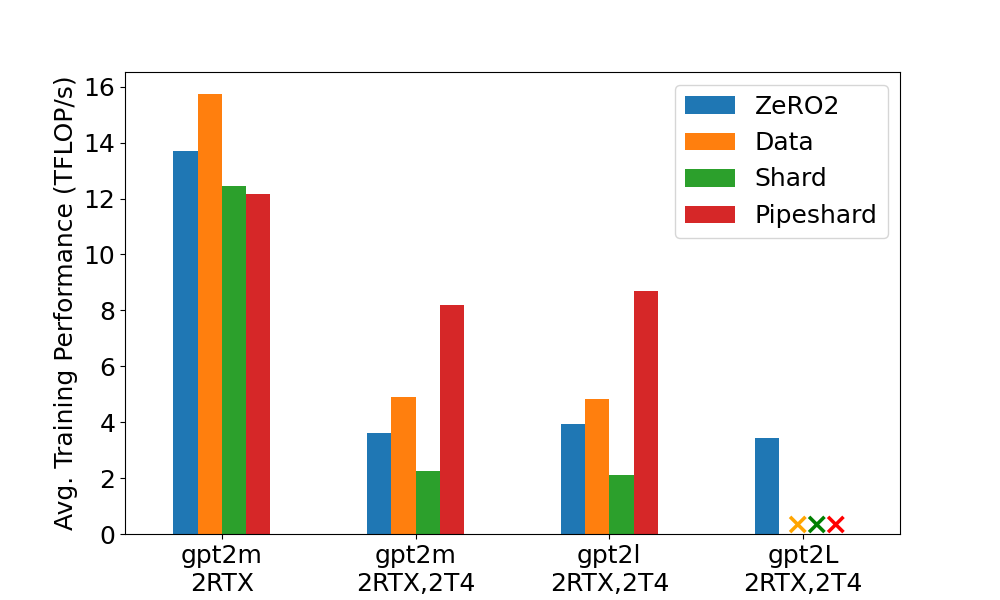} \\
(b) Training performance
\end{tabular}
\end{center}
\caption{SLM pretraining performance on TACC ($\mathbf{\times}$ denotes failed execution)}
\label{fig-tacc-tacc-results}
\end{figure}

We first report the results for TACC-TACC (single-site). Figure~\ref{fig-tacc-tacc-results} shows the time taken (for 20 epochs) and average training performance (in TFLOP/s) for different GPU clusters. When all 4 GPUs (2 RTX and 2 T4) were used for \gptm{} and \gptl, \pipeshard{} achieved the best performance due to its ability to combine intra-operator and inter-operator parallelization. Two stages were executed on two $(1\times2)$ meshes; each mesh belonged to one VM; and pipeline parallelism was employed between the two stages. \shard{} had the worst performance due to its higher overhead of GPU-GPU collective communication~\cite{zheng2022alpa} that became more pronounced due to TCP/IP being used by NCCL. For example, with \gptl, \pipeshard{} achieved 8.71 TFLOP/s compared to the second best performing approach (i.e., \data), which achieved only 4.82 TFLOP/s. When the model size increased to \gptL, however, \zero{} was the only approach that executed successfully without \textit{out-of-memory} errors. \pipeshard{} generally requires more memory for pretraining and can fail when participating GPUs are heterogenous with different GPU memory sizes.

For pretraining using only 2 RTX GPUs (in a single VM), \data{} and \zero{} performed better than the others when the models could fit within GPU memory. For example, with \gptm, \data{} achieved 15.74 TFLOP/s compared to \pipeshard's 12.17 TLFLOP/s. \pipeshard{} did not offer much benefit compared to traditional data parallelism. It is interesting that for \gptm, running on 2 RTX was faster (with \data{} and \zero) than using \pipeshard{} on 2 RTX and 2 T4. Thus, using more GPUs may not always lead to faster pretraining especially when spanning multiple VMs. The number of GPUs required for pretraining would truly depend on the model size.

For pretraining \gptL{} with 2 RTX, all techniques failed to execute due to out-of-memory errors. Hence, those results are not shown in Figure~\ref{fig-tacc-tacc-results}.


\subsection{UTAH-GPN}

\begin{figure}[h]
\begin{center}
\begin{tabular}{c}
\includegraphics[width=3.6in, angle=0]{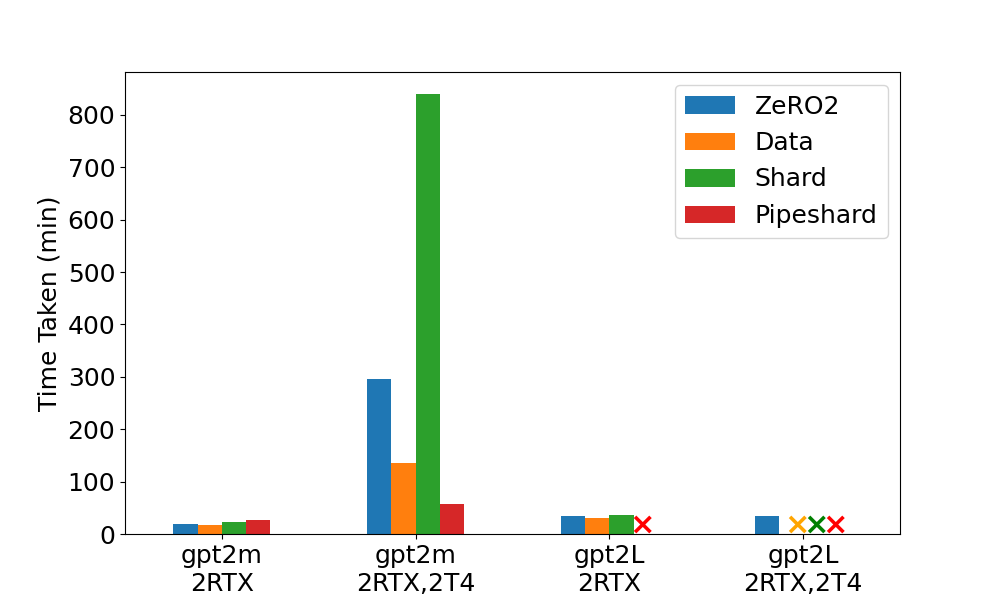} \\
(a) Execution time\\
\includegraphics[width=3.6in, angle=0]{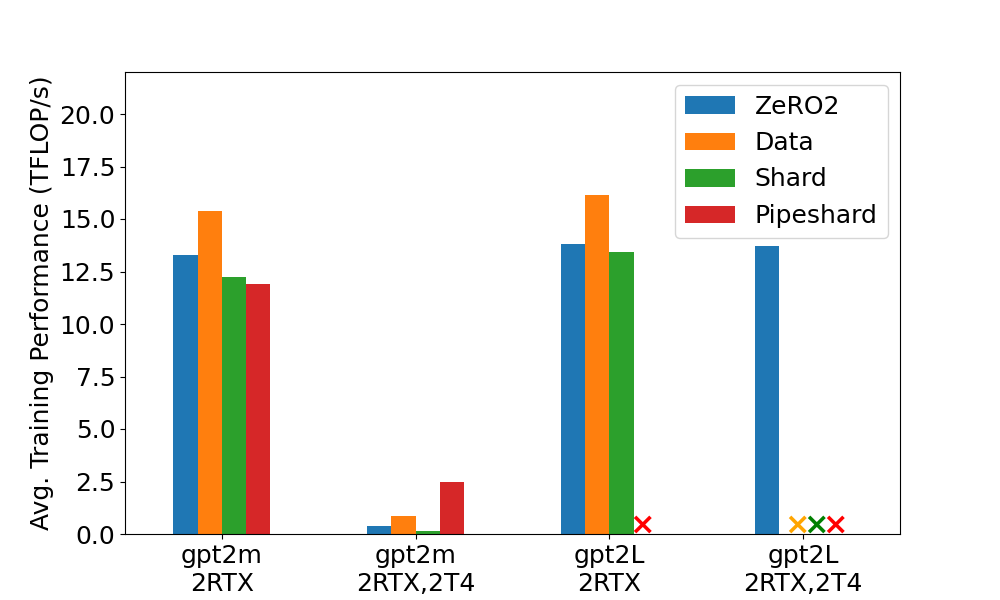} \\
(b) Training performance
\end{tabular}
\end{center}
\caption{SLM pretraining on UTAH-GPN ($\mathbf{\times}$ denotes failed execution)}
\label{fig-utah-gpn-results}
\end{figure}

Next, we report the results for UTAH-GPN (two-site). While the VM configurations and GPU hardware in UTAH-GPN were identical to that of TACC-TACC, the VMs, however, were geographically distributed with non-trivial network latency between them (see Table~\ref{table-FABRIC-slices}). Figure~\ref{fig-utah-gpn-results} reports the time taken (for 20 epochs) and the average training performance for different GPU clusters. When all 4 GPUs (2 RTX and 2 T4) were used for \gptm, \pipeshard{} achieved the best performance as before. \shard{} had the worst performance due to its higher overhead of GPU-GPU communication. For example, with \gptm, \pipeshard{} achieved 2.49 TFLOP/s compared to the second best performing approach (i.e., \data), which achieved 0.88 TLFLOP/s. The lower overall training performance can be clearly attributed to increase in network latency between the 2 VMs. For \gptL, \zero{} executed successfully while the others failed due to out-of-memory errors.

For pretraining \gptL{} using only 2 RTX GPUs (in a single VM), \data, \zero, and \shard{} executed successfully and had similar performance. Essentially, \pipeshard{} requires more GPU memory to execute than the other techniques. For \gptm, \data{} performed better than \pipeshard{} (18 min vs 26 min).

\subsection{UTAH-MASS}

\begin{figure}[h]
\begin{center}
\begin{tabular}{c}
\includegraphics[width=3.6in, angle=0]{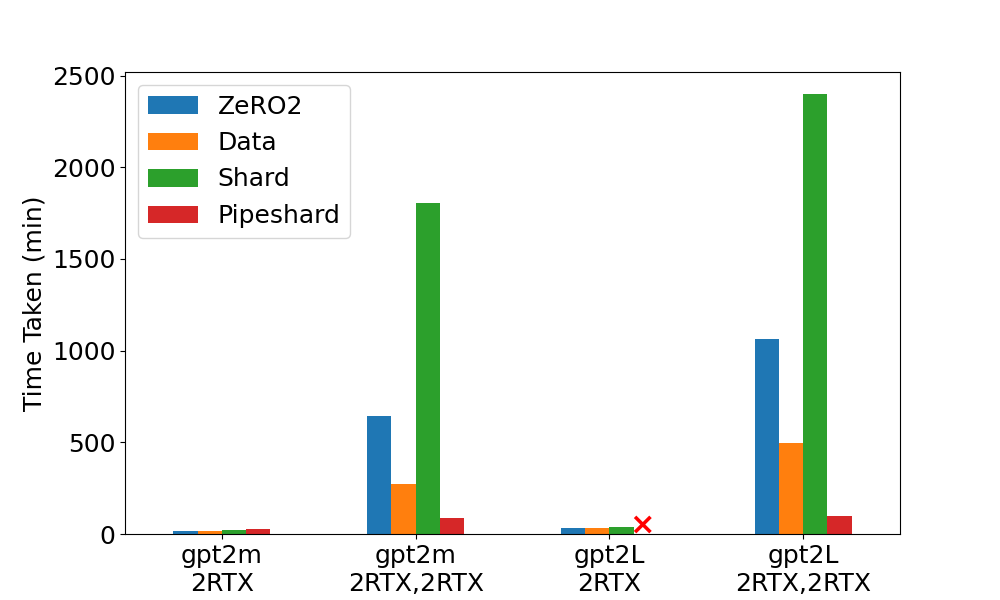} \\
(a) Execution time\\
\includegraphics[width=3.6in, angle=0]{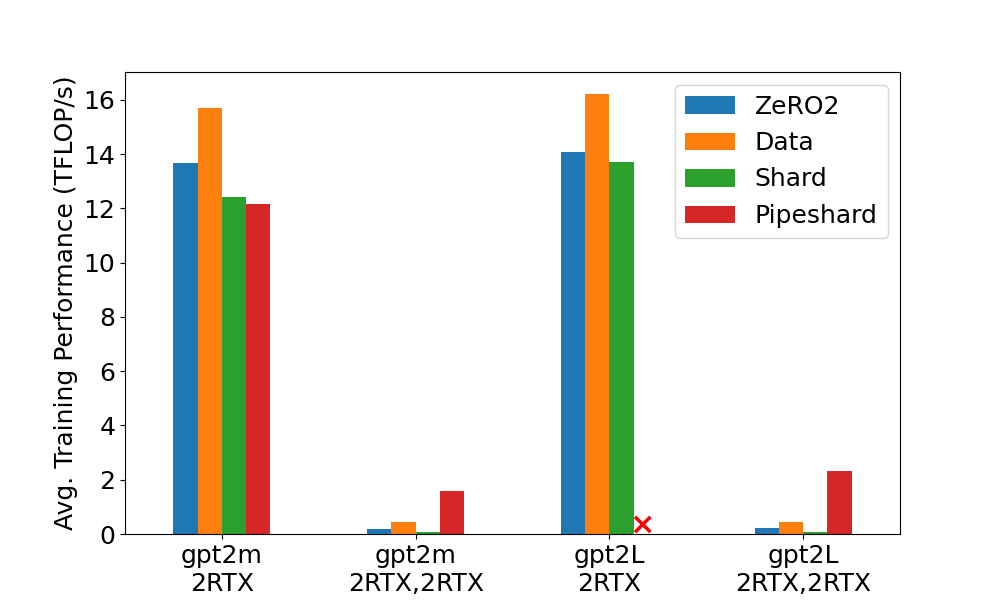} \\
(b) Training performance
\end{tabular}
\end{center}
\caption{SLM pretraining on UTAH-MASS ($\mathbf{\times}$ denotes failed execution)}
\label{fig-utah-mass-results}
\end{figure}

Next, we report the results for UTAH-MASS (two-site). This GPU cluster was similar to that of UTAH-GPN in terms of the hardware configurations; however, the network latency between the VMs was nearly three times (see Table~\ref{table-FABRIC-slices}). Figure~\ref{fig-utah-mass-results} reports the time taken (for 20 epochs) and the average training performance for different GPU clusters. When all 4 GPUs (4 RTX) were used for \gptm{} and \gptL, \pipeshard{} achieved the best performance as before. \shard{} had the worst performance. For example, with \gptL, \pipeshard{} achieved 2.32 TFLOP/s compared to the second best performing approach (i.e., \data), which achieved 0.44 TFLOP/s. The lower overall training performance can be attributed to increase in network latency between the 2 VMs.

For pretraining \gptm{} and \gptL{} using only 2 RTX GPUs (in a single VM), \data, \zero, and \shard{} executed successfully. However, \pipeshard{} failed to execute for \gptL{} due to its higher memory requirement compared to the others. We also observed that a single VM with 2 GPUs achieved better training performance compared to using 2 VMs with 4 GPUs for the tested models. Hence, picking the right number of GPUs is necessary for achieving pretraining good performance.

Compared to UTAH-GPN and TACC-TACC, UTAH-MASS had higher total GPU memory. Hence, \pipeshard{} ran successfully for \gptL{} using 4 RTX GPUs.

\subsection{BRIS-STAR}

\begin{figure}[h]
\begin{center}
\begin{tabular}{c}
\includegraphics[width=3.6in, angle=0]{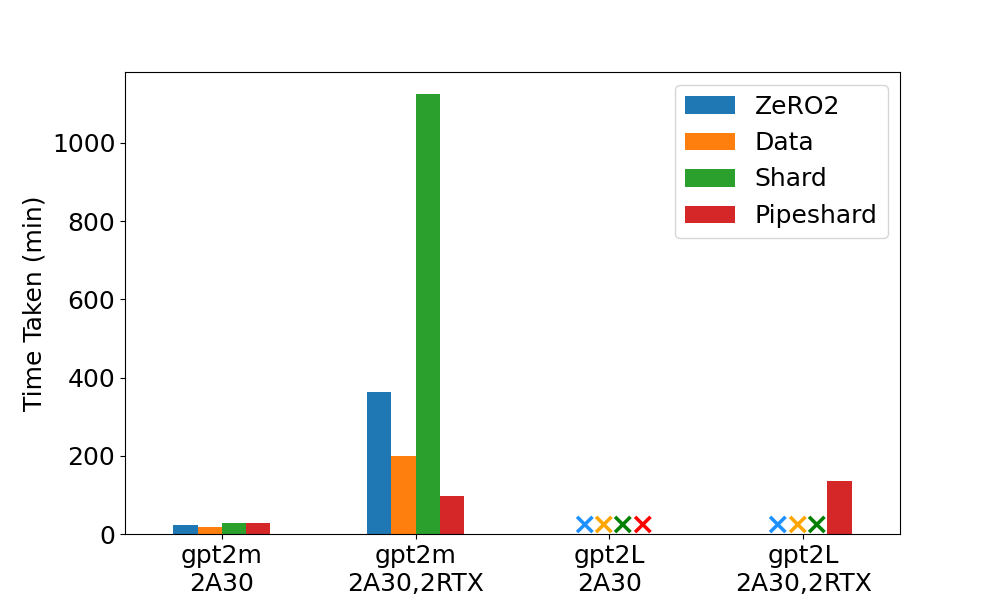} \\
(a) Execution time\\
\includegraphics[width=3.6in, angle=0]{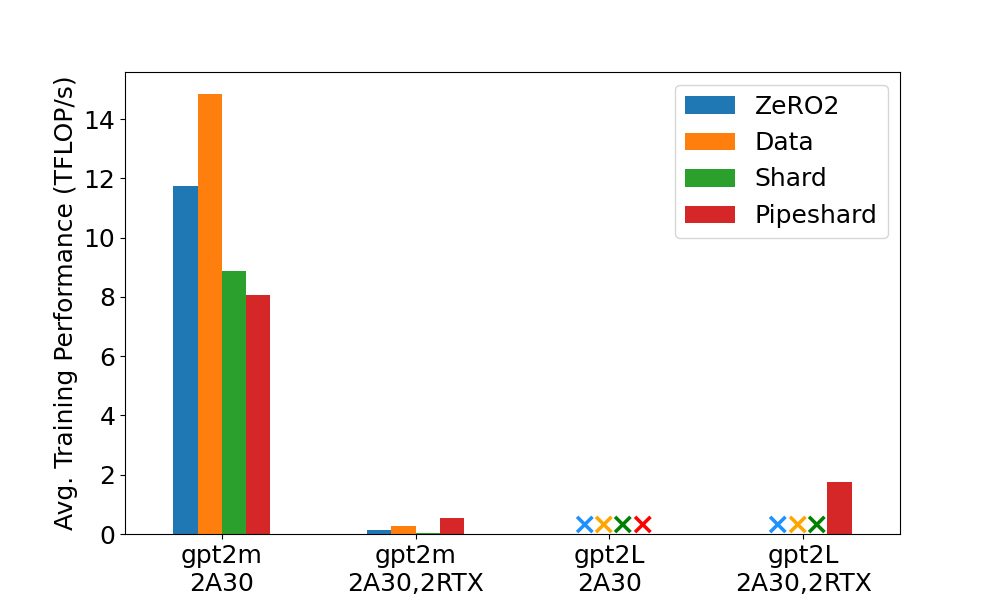} \\
(b) Training performance
\end{tabular}
\end{center}
\caption{SLM pretraining on BRIS-STAR ($\mathbf{\times}$ denotes failed execution)}
\label{fig-bris-star-results}
\end{figure}

Next, we report the results for BRIS-STAR (two-site). Compared to UTAH-MASS, this two-site setting had heterogenous GPUs (2 A30 and 2 RTX) and even higher network latency between the VMs (see Table~\ref{table-FABRIC-slices}). Figure~\ref{fig-bris-star-results} reports the time taken (for 20 epochs) and the average training performance for different GPU clusters. When all 4 GPUs (2 A30 and 2 RTX) were used for \gptm{} and \gptL, \pipeshard{} achieved the best performance as before. \shard{} had the worst performance. For example, with \gptL, \pipeshard{} achieved 1.77 TFLOP/s while other approaches failed to execute due to out-of-memory errors. With \gptm, \pipeshard's training performance was twice that of \data{} (0.55 TFLOP/s vs 0.26 TFLOP/s). 


For pretraining \gptm{} with only 2 A30 (single-site), all approaches executed successfully. \data{} achieved the best performance. We also observed better performance for \gptm{} with a single VM than with 2 VMs. However, for \gptL, \pipeshard{} on all 4 GPUs was the only successful execution.

\subsection{GAT-AMST}

\begin{figure}[h]
\begin{center}
\begin{tabular}{c}
\includegraphics[width=3.6in, angle=0]{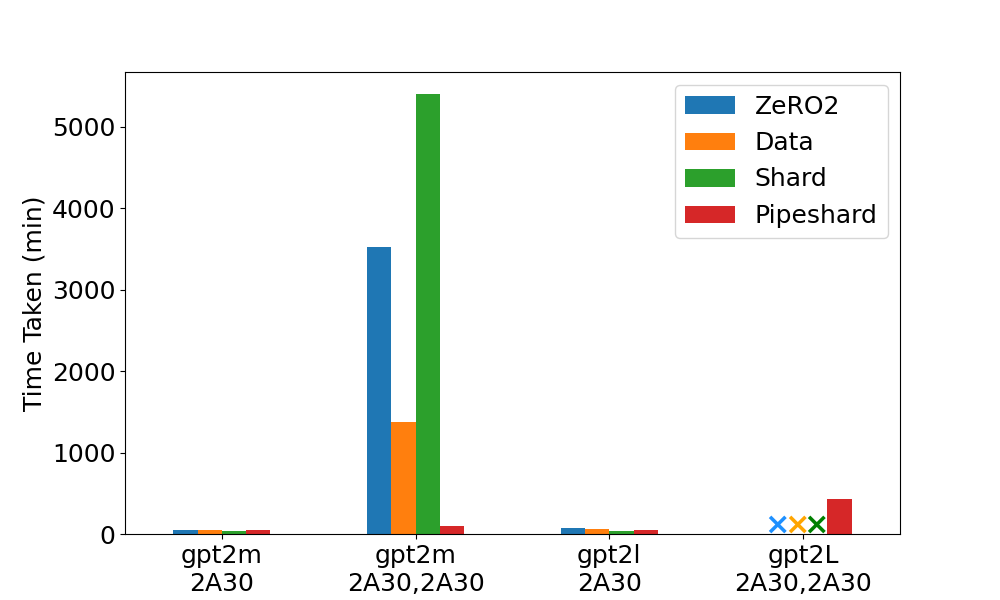} \\
(a) Execution time\\
\includegraphics[width=3.6in, angle=0]{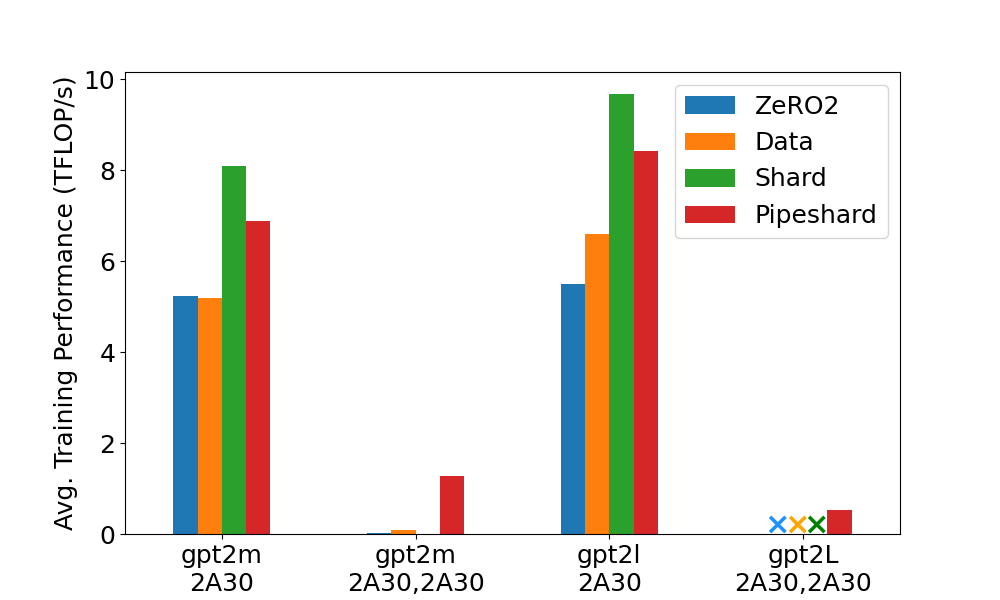} \\
(b) Training performance
\end{tabular}
\end{center}
\caption{SLM pretraining on GAT-AMST ($\mathbf{\times}$ denotes failed execution)}
\label{fig-gat-amst-results}
\end{figure}

Finally, we report the results for GAT-AMST (two-site). This experiment had homogeneous GPU hardware (4 A30) but with the worst network latency among all the experiments (see Table~\ref{table-FABRIC-slices}). Figure~\ref{fig-gat-amst-results} reports the time taken (for 20 epochs) and the average training performance for different GPU clusters. When all 4 A30 GPUs were used for \gptm{} and \gptL, \pipeshard{} achieved the best performance as before. In fact, \data, \zero, and \shard{} failed to execute for \gptL. \pipeshard{} achieved 0.52 TFLOP/s and 1.28 TFLOP/s for \gptL{} and \gptm, respectively.

However, using a single VM (and 2 A30), for \gptL, all approaches failed to execute due to out-of-memory errors similar to BRIS-STAR. Hence, it is not shown in Figure~\ref{fig-gat-amst-results}. So we tested with \gptl, a smaller model than \gptL. All the techniques executed successfully with \shard{} achieving the best performance of 9.68 TFLOP/s. It required 41 min compared to 52 min required by \pipeshard. For \gptm, similar trend was observed. \shard{} executed in 34 min compared to 44 min required by \pipeshard. It is interesting that \shard{} performed better than \data{} on 2 A30 GPUs compared to on 2 RTX GPUs (in TACC-TACC and UTAH-GPN). This can be attributed to \textit{faster memory bandwidth} of A30 compared to RTX.

\subsection{Impact of Network Latency}

We also report the impact of network latency on SLM pretraining performance. Table~\ref{table-pretraining-latency} shows the time taken by different techniques in different two-site GPU clusters. As observed, increase in network latency deteriorates the training performance, with \shard{} being affected the most, due to increased overhead of GPU-GPU collective communication. On the other hand, \pipeshard{} tolerated the increased latencies much better due to point-to-point communication between VMs for the two stages created by inter-operator/pipeline parallelism. Compared to \data, \zero{} suffered higher performance degradation due to increase in network latency.

\begin{table}[t]
\begin{center}
\caption{Pretraining performance for \gptm. The experiments are ordered by increasing site-site network latency between the 2 VMs. Best timings are shown in bold.}
\label{table-pretraining-latency}
\begin{tabular}{|c|c|c|c|c|c|}
\hline
\multicolumn{1}{|c|}{} &
\multicolumn{5}{|c|}{\textbf{Time Taken (min) for 20 Epochs}} \\
\cline{2-6}
\multicolumn{1}{|c|}{\textbf{Technique}} &
\multicolumn{1}{|c|}{TACC-} &
\multicolumn{1}{|c|}{UTAH-} &
\multicolumn{1}{|c|}{UTAH-} &
\multicolumn{1}{|c|}{BRIS-} &
\multicolumn{1}{|c|}{GAT-} \\
\multicolumn{1}{|c|}{} &
\multicolumn{1}{|c|}{TACC} &
\multicolumn{1}{|c|}{GPN} &
\multicolumn{1}{|c|}{MASS} &
\multicolumn{1}{|c|}{STAR} &
\multicolumn{1}{|c|}{AMST} \\
\hline
\data & 41 & 136 & 272 & 199 & 1,375 \\
\hline
\zero & 52 & 295 & 641 & 363 & 3,519 \\
\hline
\shard & 82 & 840 & 1,808 & 1,125 & 5,400\\
\hline
\pipeshard & \textbf{29} & \textbf{57} & \textbf{86} & \textbf{96} & \textbf{100} \\
\hline
\end{tabular}    
\end{center}
\end{table}

\subsection{Summary of Experimental Results}

Based on the results of our evaluation, we summarize the following observations:
\begin{enumerate}
\item In a two-site GPU cluster, \pipeshard{} achieved the best training performance for an SLM. With increasing site-site latencies, the performance of \data, \shard, and \zero{} deteriorated faster. However, \pipeshard{} tolerated increase in network latencies much better than the others.
\item When pretraining with a single-site GPU cluster executed successfully for \zero, \data, or \shard, they achieved better performance than \pipeshard{} in the same setting. Out of 8 experiments, \data{} was the winner in six, and \shard{} was the winner in two. 
\item \pipeshard{} typically required more memory compared to the others. When \pipeshard{} failed in a two-site GPU cluster due to memory issues (e.g., when using heterogenous GPU hardware), \zero{} was able to execute successfully due to its lower memory utilization.
\end{enumerate}

Thus, a user must be cautious in selecting a particular pretraining technique and a GPU cluster configuration to achieve good SLM pretraining performance. 

\subsection{Pretraining Technique Selection}

We expect to pretrain SLMs for hundreds of epochs on input datasets to obtain low perplexity scores during evaluation. (This will also enable more accurate embeddings on raw data to build vector databases for downstream AI applications.) Therefore, pretraining may last for several days to a week or more. Hence, we must select an appropriate pretraining technique given an input SLM and a GPU cluster configuration. Without loss of generality, suppose there are two VMs with $2^{n}$ GPUs ($n \geq 0$) attached to each VM. The VMs may be located in a single site or span across two sites. The below algorithm sketch (see Algorithm~\ref{algo-pretraining-technique}) describes how to select an appropriate pretraining technique to achieve \textit{high} training performance and \textit{lower} total training time as well as to \textit{reduce} the number of GPUs used.

The main steps of the algorithm are as follows: We first run \pipeshard{} on both VMs using all the GPUs and measure the avg. training performance for a small number of epochs $\epsilon$ (Lines~\ref{algo-pipeshard-2VM-start}-\ref{algo-pipeshard-2VM-end}). We then execute \data{} and \shard{} for $\epsilon$ epochs on each VM separately and measure the avg. training performance similarly (Lines~\ref{algo-data-shard-start}-\ref{algo-data-shard-end}). If \pipeshard{} on 2 VMs has better training performance than the best performance on 1 VM (using \data{} or \shard), then \pipeshard{} is selected to run using both VMs (Lines~\ref{algo-pipeshard-start}-\ref{algo-pipeshard-end}). We allow the user to specify a threshold of $\delta$ for controlling the desired training performance improvement. Otherwise, \data{} or \shard{} on 1 VM is a better choice (Lines~\ref{algo-data-shard-start}-\ref{algo-data-shard-end}). If none of the above conditions are satisfied, then \zero{} is executed on all the GPUs using both VMs. If successful, \zero{} is the preferred technique. Otherwise, a larger cluster with more GPU memory is needed for pretraining. These steps are shown in Lines~\ref{algo-zero-start}-\ref{algo-zero-end}.  

\begin{algorithm}[p]
\caption{Pretraining technique selection}
\label{algo-pretraining-technique}
\begin{algorithmic}[1]
\REQUIRE $\mathbb{M}$ - SLM to train; $\mathbb{V}_1$ - A VM with GPUs in one site; $\mathbb{V}_2$ - A VM with GPUs in another site; $\delta \in (0,1]$ - a user-specified threshold; $\epsilon$ - small number of epochs
\ENSURE Best pretraining technique and VMs for pretraining
\STATE Pretrain $\mathbb{M}$ using \pipeshard{} for $\epsilon$ epochs on $\mathbb{V}_1 \cup \mathbb{V}_2$\label{algo-pipeshard-2VM-start}\COMMENT{Check \pipeshard's performance on the cluster}
\STATE Let $T_p$ denote the avg. training performance (e.g., TFLOP/s) over $\epsilon$ epochs; set $T_p \leftarrow 0$ if failure occurs\label{algo-pipeshard-2VM-end}
\STATE Pretrain $\mathbb{M}$ using \data{} for $\epsilon$ epochs on $\mathbb{V}_1$\label{algo-data-shard-start} \COMMENT{Check \data's performance on the first VM}
\STATE Let $T_{d1}$ denote the avg. training performance on $\mathbb{V}_1$ over $\epsilon$ epochs; set $T_{d1} \leftarrow 0$ if failure occurs
\STATE Pretrain $\mathbb{M}$ using \shard{} for $\epsilon$ epochs on $\mathbb{V}_1$ \COMMENT{Check \shard's performance on the first VM}
\STATE Let $T_{s1}$ denote the avg. training performance on $\mathbb{V}_1$ over $\epsilon$ epochs; set $T_{s1} \leftarrow 0$ if failure occurs
\STATE Pretrain $\mathbb{M}$ using \data{} for $\epsilon$ epochs on $\mathbb{V}_2$ \COMMENT{Check \data's performance on the second VM}
\STATE Let $T_{d2}$ denote the avg. training performance on $\mathbb{V}_2$ over $\epsilon$ epochs; set $T_{d2} \leftarrow 0$ if failure occurs
\STATE Pretrain $\mathbb{M}$ using \shard{} for $\epsilon$ epochs on $\mathbb{V}_2$ \COMMENT{Check \shard's performance on the second VM}
\STATE Let $T_{s2}$ denote the avg. training performance on $\mathbb{V}_2$ over $\epsilon$ epochs; set $T_{s2} \leftarrow 0$ if failure occurs\label{algo-data-shard-end}
\STATE $T_z = \max(T_{d1}, T_{d2}, T_{s1}, T_{s2})$\label{algo-pipeshard-start} \COMMENT{Choose the best training performance}
\IF{$T_z > 0$ and $\frac{T_p-T_z}{T_z} > \delta$} 
\RETURN \pipeshard, \{$\mathbb{V}_1 \cup \mathbb{V}_2$\}\label{algo-pipeshard-end} \COMMENT{\pipeshard{} is a better choice on the entire cluster}
\ELSIF{$T_p > 0$ and $\frac{T_z-T_p}{T_p} > \delta$}\label{algo-data-shard-start} 
\IF{$\max(T_{d1}, T_{s1}) \geq \max(T_{d2}, T_{s2})$}
\IF{$T_{d1} \geq T_{s1}$}
\RETURN \data, \{$\mathbb{V}_1$\} \COMMENT{\data{} is a better choice}
\ELSE
\RETURN \shard, \{$\mathbb{V}_1$\} \COMMENT{\shard{} is a better choice}
\ENDIF
\ELSE
\IF{$T_{d2} \geq T_{s2}$}
\RETURN \data, \{$\mathbb{V}_2$\} \COMMENT{\data{} is a better choice}
\ELSE
\RETURN \shard, \{$\mathbb{V}_2$\} \COMMENT{\shard{} is a better choice}
\ENDIF
\ENDIF \label{algo-data-shard-end}
\ELSE
\STATE Pretrain $\mathbb{M}$ using \zero{} for $\epsilon$ epochs on $\mathbb{V}_1 \cup \mathbb{V}_2$\label{algo-zero-start} \COMMENT{Check if \zero{} works on the entire cluster}
\STATE Let $T_{z'}$ denote the avg. training performance over $\epsilon$ epochs; set $T_{z'} \leftarrow 0$ if failure occurs
\IF{$T_{z'} > 0$}
\RETURN \zero, \{$\mathbb{V}_1 \cup \mathbb{V}_2$\} \COMMENT{\zero{} is a better choice on the entire cluster}
\ELSE
\RETURN $\emptyset, \emptyset$ \COMMENT{Need more GPU memory}
\ENDIF \label{algo-zero-end}
\ENDIF
\end{algorithmic}
\end{algorithm}






%% file: conclusion.tex
\section{Conclusion}
\label{sec-conclusion}

We presented an empirical study of different SLM pretraining techniques on FABRIC, a nationwide research infrastructure available for academic research at no charge. We evaluated \data, \zero2, \shard, and \pipeshard{} for pretraining GPT-2 models on GPU clusters that either spanned a single-site or across two-sites on FABRIC. \pipeshard, that collectively optimized both intra-operator parallelism and inter-operator/pipeline parallelism, consistently achieved better training performance than other techniques especially when GPUs were geographically distributed. The performance of \data, \zero, and \shard{} deteriorated with increase in network latency due to the communication overhead. In a single-site setting, \data{} typically achieved better training performance when the models fit in GPU memory. Finally, we proposed a systematic approach for selecting the appropriate SLM pretraining technique to achieve high training performance/lower total execution time and lower the number of GPUs used. Although our experiments were conducted on FABRIC, the insights gained can be applied to other infrastructures that have similar GPU cluster configurations (e.g., campus computing, experimental testbeds). We believe by democratizing SLM pretraining on user-specified datasets, more accurate embeddings of raw data can be generated for domain-specific vector databases to support downstream AI applications. 


\section*{Acknowledgments}

This work was supported by the National Science Foundation under Grant No. 2502893.